%% file: ms.tex
\documentclass[12pt,a4paper,final]{report}
\usepackage[utf8]{inputenc}
\usepackage[english]{babel}
\usepackage{amsmath}
\usepackage{amsfonts}
\usepackage{amssymb}
\usepackage{graphicx}
\usepackage{watermark}
\usepackage{transparent}
\usepackage{siunitx}
\usepackage[square,sort,comma,numbers,numbers,sort&compress]{natbib}
\usepackage[hidelinks]{hyperref}
\usepackage{ragged2e}
\usepackage[font=small,labelfont=bf]{caption}
\addto\captionsenglish{}
\addto\captionsenglish{}
\usepackage{listings}  
\usepackage[nottoc]{tocbibind}

\begin{document}

\pagenumbering{gobble}
\pagenumbering{roman}
\include{pages/abs_arxiv}
\tableofcontents
\listoftables
\listoffigures
\include{pages/symbol}
\break
\pagenumbering{arabic}
\include{pages/chap1}

\include{pages/chap2}

\include{pages/chap3}

\include{pages/chap}
\include{pages/chap4}

\include{pages/chap5}

\bibliography{bib/ms}
\bibliographystyle{apa-good}
\end{document}

%% file: pages/abs_arxiv.tex
 
\begin{center}
\begin{Large}  \textbf {MAKING NEURAL MACHINE READING COMPREHENSION FASTER}
	\end{Large}	 
       \vspace{0.5cm}
       \linespread{1.5}\selectfont
       \newline
       \textit{
       	This study aims at solving the Machine Reading Comprehension problem where questions have to be answered given a context passage. The challenge is to develop a computationally faster model which will have improved inference time. State of the art in many natural language understanding tasks, BERT model, has been used and knowledge distillation method has been applied to train two smaller models. The developed models are compared with other models which have been developed with the same intention.
       }
       
       \vspace{4.5cm}

\end{center}

%% file: pages/symbol.tex
\begin{chapter}*{Glossary}
\addcontentsline{toc}{chapter}{Glossary}

\begin{tabular}
{cp{0.5\textwidth}}
  MRC & Machine Reading Comprehension \\    
  \\
  NLP & Natural Language Processing\\
  \\
  SQuAD & Stanford Question Answering Dataset \\
  \\
  RNN & Recurrent Neural Network \\
  \\
  CNN & Convolution Neural Network\\
  \\
  SOTA & State of the Art   
\end{tabular}

\end{chapter}

%% file: pages/chap1.tex
\begin{chapter} {Introduction}

	Machine Reading Comprehension is one of the key problems in Natural Language Understanding, where the task is to read and comprehend a given text passage, and then answer questions based on it. This task is challenging which requires a comprehensive understanding of natural language and the ability to do further inference and reasoning on top of it.  Stanford Question Answering Dataset (SQuAD)~\cite{DBLP:journals/corr/RajpurkarZLL16}, which is used in this study, is a reading comprehension dataset introduced by Rajpurkar et al. (2016) and contains over 100,000 question-answer pairs on over 500 Wikipedia articles. Each question-answer pair contains a question with a correct answer that is a span of text from the corresponding reading passage (context).
	\begin{center}
	\textit{\textbf{Question : }}\textit{What company owns the American Broadcasting Company?} \linebreak 
	\textit{\textbf{Context : }}\textit{The American Broadcasting Company (ABC) (stylized in its logo as abc since 1957) is an American commercial broadcast television network that is owned by the Disney–ABC Television Group, a subsidiary of Disney Media Networks division of \textbf{The Walt Disney Company}. The network is part of the Big Three television networks. The network is headquartered on Columbus Avenue and West 66th Street in Manhattan, with additional major offices and production facilities in New York City, Los Angeles and Burbank, California.
	} \linebreak
	\textit{\textbf{Answer : }}\textit{The Walt Disney Company} 
	\end{center}
	Over the past few years, significant progress has been made with end-to-end models showing promising results on many challenging datasets. The backbone of the majority of models contain two key ingredients: 
	\begin{itemize}
		\item A word level recurrent model to process sequential inputs. In this case, the inputs are question and context text.
		\item An attention mechanism for long term interactions between those sequential inputs.
		
	\end{itemize}
	However, these models, because of their recurrent architecture, face slow training and inference time. This becomes worse especially when the reading passage is very long, as the recurrent models process words sequentially. Because of this, although these models have been producing promising results, utilizing these models in a real life environment becomes impossible. \linebreak \linebreak
	To cope up with this problem, RNN free architectures like QANet~\cite{DBLP:journals/corr/abs-1804-09541}, which combines local convolution over words with a global self-attention mechanism, has been developed. In this study, BERT model~\cite{DBLP:journals/corr/abs-1810-04805} has been used with the intention of building a computationally faster neural architecture which performs reasonably well on SQuAD task.
\end{chapter}

%% file: pages/chap2.tex
\begin{chapter} {Literature Review}
	Most of the papers that approach to solve SQuAD problem have utilized RNN based models ~\citep{
		DBLP:journals/corr/YuZHYXZ16,
		DBLP:journals/corr/HuPQ17,
		DBLP:journals/corr/PanLZCCH17,
		DBLP:journals/corr/abs-1712-03609,
		DBLP:journals/corr/abs-1712-03556,
		mcr, 
		DBLP:journals/corr/abs-1711-07341, DBLP:journals/corr/abs-1711-00106, DBLP:journals/corr/abs-1710-10504, DBLP:journals/corr/abs-1710-10723, DBLP:journals/corr/abs-1710-02772, DBLP:journals/corr/GongB17, DBLP:journals/corr/ZhangZCDWJ17, DBLP:journals/corr/ShenHGC16, DBLP:journals/corr/ChenFWB17, DBLP:journals/corr/LeeKP016, DBLP:journals/corr/WeissenbornWS17, DBLP:journals/corr/WangMHF16, DBLP:journals/corr/LiuHWYN17, DBLP:journals/corr/SeoKFH16, DBLP:journals/corr/XiongZS16, DBLP:journals/corr/WangJ16a, DBLP:journals/corr/YangDYHCS16, DBLP:journals/corr/BahdanauBJGVB17} and they follow a similar chain of processes. This begins with pre-trained word-embeddings that are then processed by bidirectional RNNs. Question and context are processed independently, and their interaction is modeled by attention mechanisms to produce an answer. There are small differences in the type of attention each model applies, but in every model, it is calculated over the hidden states of an RNN. \\ \\
	Vaswani et al.~\cite{DBLP:journals/corr/VaswaniSPUJGKP17} applied attention directly over the word-embeddings, and derived a new neural network architecture, Transformer, which achieved state-of-the-art results in machine translation without any RNN. Scaled multi-headed dot product attention mechanism, proposed by Vaswani et al., has been heavily used in RNN free neural architectures like QANet, BERT, GPT~\cite{radford2019language} for language modeling task.  In this study, the same attention mechanism has been used.
	To eliminate the need for a recurrent structure in the architecture of the network, various studies have employed different kinds of CNN, namely depthwise separable CNN in QANet. FABIR~\cite{DBLP:journals/corr/abs-1810-09580} uses a convolutional attention method. Bell et al.~\cite{DBLP:journals/corr/abs-1810-08680} also included CNN in their model, in their search for a faster neural architecture for MRC problem.
	\begin{section}{Attention Mechanism}
	Attention Mechanisms are used to model interactions between elements of different input sequences or a single sequence. It has been successfully applied in models aimed to solve various tasks of NLP, like machine translation\cite{DBLP:journals/corr/abs-1812-07807, DBLP:journals/corr/VaswaniSPUJGKP17, DBLP:journals/corr/abs-1805-04237}  and natural language inference task\cite{DBLP:journals/corr/LiuQH16b, DBLP:journals/corr/abs-1708-01353}.
	\\ \\
	Let’s assume we have two sets of word vectors $P = \{p_1, ..., p_m\}$ and $Q = \{q_1,..., q_n\}$. The first building block of the attention mechanism is the score function which gives a scalar score $\alpha_{ij}$ to $p_i\in P$ with respect to $q_j\in Q$. Mathematically this can be expressed as,
	\begin{gather}
	s_{ij} = f(p_i, q_j)\\
	\alpha_{ij} = \frac{e^{s_{ij}}}{\sum_{k=1}^{n}e^{s_{ik}}}
	\end{gather}
	Now this score $\alpha_{ij}$ can be used to model interactions from elements of $P$ to $Q$, which can be done by taking weighted sum of all vectors in $Q$ for all elements in $P$.
	\begin{gather}
	c_i = \sum_{j=0}^{n}\alpha_{ij}q_j
	\end{gather}
	Here, $C = \{c_i, ..., c_m\}$ reflects the interaction from $P$ to $Q$. In various studies, different kinds of $f$ have been proposed. Some examples are given below,
	\begin{gather}
	f(p_i, q_j)=
	\begin{cases}
	p_i^TU^TVq_j & \text{multiplicative}\\
	\frac{p_i^TU^TVq_j}{\sqrt{k}} & \text{scaled multiplicative}\\
	v^Ttanh(W^2p_i + W^3q_j) & \text{additive}\\	
	\end{cases}
	\end{gather}
	In the above examples all notations other than $p_i$, $q_j$ and $k$ are variable and trainable. $k$ is the attention hidden size.
	\end{section}
	\begin{section}{BERT\textsubscript{BASE}}
	The architecture of the BERT\textsubscript{BASE} model\cite{DBLP:journals/corr/abs-1810-04805}, which is extensively used in this study, can be described as a multi-layer bidirectional Transformer encoder. BERT\textsubscript{BASE} architecture has a embedding layer followed by 12 Transformer encoder layers\cite{DBLP:journals/corr/VaswaniSPUJGKP17}, with hidden size 768 and has 110 million parameters. The model was trained on 2 types of pre-training tasks, namely, Masked Language Modeling and Next Sentence Prediction, which is well explained in the paper\cite{DBLP:journals/corr/abs-1810-04805}. For SQuAD task, A span prediction layer is appended at the end of the model, which is explained in Section 3.1.2. After fine-tuning for SQuAD task, BERT\textsubscript{BASE} archives SOTA accuracy in many NLP tasks.  
	\end{section}
\end{chapter}

%% file: pages/chap3.tex
\begin{chapter} {Model Description}
		\begin{section}
			{Conv Model}
			The model consists of four main building blocks:
			\begin{enumerate}
				\item Embedding Layer
				\item Multi-window Convolution Layer
				\item Transformer Encoder Layer
				\item Span Prediction Layer
								
			\end{enumerate}
		\begin{subsection}{Embedding Layer}
			The embedding layer takes the context and question text and creates a vectorized input representation, for each context-question pair. WordPiece tokenization (Wu et al., 2016) has been used with a 30,000 token vocabulary. Question-context pairs with questions consisting more than 60 tokens are discarded. After tokenizing the question and context text, the input representation with maximum sequence length 384, excluding the 10 padding tokens between the last question token and SEP token, looks like below:
	\begin{figure}
			\begin{center}
				\resizebox{\textwidth}{!}{%
				\begin{tabular}{|p{0.7in}|p{0.7in}|p{0.8in}|p{0.7in}|p{0.7in}|p{0.7in}|p{0.7in}|}
					\hline
					NO\_ANS token & Question tokens & 10 Padding tokens & SEP token & Context tokens & SEP token & Padding tokens
					 \\  
					 \hline
				\end{tabular}}
			\end{center}
	\caption{Input Representation}
	\label{faketable:mul}
	\end{figure}
\linebreak \linebreak
			If the number of context tokens is more than (384 - the number of question tokens - 3) , a sliding window approach is used where chunks of the Context tokens are taken with a stride of 128. As the answer span won’t be used in every chunk of a context text, NO\_ANS token is used in this scenario during span prediction, detailed explanation follows in the Training section. For a given token, its input representation is constructed by summing the corresponding token, segment and position embeddings, followed by a layer-normalization\cite{DBLP:journals/corr/BaKH16} layer. For positional and segment embeddings, 384 and 2 vectors are used respectively. Here, the question tokens are set as segment 1 and context tokens are set as segment 2. Token and other two types of embedding vectors have been initialized from the pre-trained BERT\textsubscript{BASE} model. All the embedding vectors belong to $\mathbb{R}^{768}$. Only positional and segment tokens are trained in this layer. The final output from this layer is $\omega_i$ $\in$ $\mathbb{R}^{768}$ for each $i^{th}$ token in the sequence.
			
		\end{subsection}
	\begin{subsection}
		{Multi-window Convolution Layer
		}
	The multi-window convolution layer contains two one-dimensional depthwise separable convolutional neural networks (CNN)\cite{DBLP:journals/corr/Chollet16a}. As observed by Yu et al.\cite{DBLP:journals/corr/abs-1804-09541}, it is more memory efficient and has better generalization capability than its traditional counterpart. Each CNN is followed by a batch normalization\cite{DBLP:journals/corr/IoffeS15} layer. Leaky-RELU activation function, with $\alpha = 0.2$, has been applied elementwise, just after the CNN layer. It must be noted that the proposed model does not process the context and the question tokens separately. Both CNNs have different kernel size. To capture both the local and global view of the question and context, kernel sizes 3 and 7 have been used. All the CNN layers have $stride = 1$ and $number\text{ }of\text{ }filters =64$. At the end outputs of all the four networks are concatenated together to create a single vector for each token. The output of the later is noted as $\varphi_i \in \mathbb{R}^{128}$.
	
		\end{subsection}
	\begin{subsection}
		{Transformer Encoder Layer}
		The Transformer architecture, proposed by Vaswani et al.\cite{DBLP:journals/corr/VaswaniSPUJGKP17}, has been vastly used in Language Modeling task. In fact, it’s the only major building block of BERT and GPT. Both of the architectures, accompanied by their pre-training tasks, are considered and proven to be state of the art in many natural language tasks after fine-tuning. This layer receives input, $\varphi_i$, from the multi-window convolution layer and produces an output noted as $\psi_i \in \mathbb{R}^{128}$. The typical mathematical flow of this layer is given below.
		\begin{gather}	
		\psi_i = LayerNorm(\Lambda_i + \Gamma_i)\\
		\Gamma_i = Gelu(\Lambda_iW^I+b^I)W^O+b^O,\\
		W^I\in\mathbb{R}^{128\times256},
		W^O\in\mathbb{R}^{256\times128}, b^I\in\mathbb{R}^{256}, b^O\in\mathbb{R}^{128}\nonumber\\
		\varPhi = [\varphi_1, ..., \varphi_{384}]^T\\	
		\Lambda_i=LayerNorm(\varphi_i+MultiHead(\varPhi, \varPhi, \varPhi)_i)\\
		MultiHead(\varPhi, \varPhi, \varPhi) = [head_{(384\times{\frac{128}{H}})}^1,...,head_{(384\times{\frac{128}{H}})}^H]_{{(384\times128)}}W^A+b^A\\
		W^A\in\mathbb{R}^{128\times128}, b^A\in\mathbb{R}^{128}\nonumber\\
		q_j = \varPhi W_j^Q+[b^Q, ...,b^Q]_{(384\times\frac{128}{H})}^T\\
		k_j = \varPhi W_j^K+[b^K, ...,b^K]_{(384\times\frac{128}{H})}^T\\
		v_j =\varPhi W_j^V+[b^V, ...,b^V]_{(384\times\frac{128}{H})}^T\\
		head_j=Attention(q_j, k_j, v_j), \\
		\forall j=1,...,H; W_j^V, W_j^Q, W_j^K\in\mathbb{R}^{128\times\frac{128}{H}}; b^Q, b^V, b^K\in \mathbb{R}^{\frac{128}{H}}\nonumber\\
		Attention(q_j, k_j, v_j) = SoftMax\left(\frac{q_jk_j^T}{\sqrt{\frac{128}{H}}}\right)v_j\\
		Gelu(x)=x\times0.5\times(1.0+erf\left(\frac{x}{\sqrt{2}}\right)), \text{ }erf\text{ is Gaussian error function.}
		\end{gather}
		Here, all the $W$ and $b$ variables are trainable. Here the constant $H$ is the number of heads and $H = 4$ has been used in this study. It must be noted that all the $W$ and $b$ variables are shared among all the positions in the sequence. Here $LayerNorm$ is the layer normalization method, proposed by Ba et al\cite{DBLP:journals/corr/BaKH16}. As proposed by Devlin et al.\cite{DBLP:journals/corr/abs-1810-04805} $Gelu$ activation function is used in the hidden layer.
		
		\end{subsection}
	\begin{subsection}{Span Prediction Layer
		}
	Span prediction layer consists of a point-to-point linear layer, followed by a softmax layer. Mathematically this can be expressed as,
	\begin{gather}
	z_i = W^S\psi_i+b^S\\
	\rho_i=\frac{e^{z_i}}{\sum_{k=1}^{384}e^{z_k}}\\
	W^S\in\mathbb{R}^{2\times128}, b^S\in\mathbb{R}^2\nonumber
	\end{gather}
	10 padding tokens, which were inserted between the last question token and the SEP token in the embedding layer, are disregarded in this layer and do not take part in the computation. So the end sequence only has vectors for 384 tokens. $\rho_i$ and $z_i$ are two-dimensional vectors, where the first and second component of the vector denotes the probability and likelihood of $i^{th}$ token being the start and end token of the ground truth answer span respectively.\linebreak \linebreak
	The start and end index of the final predicted answer span is then defined by,
	\begin{equation}
	start, end = \arg\max_{i\leq j}(z_{i0}+z_{j1})
	\end{equation}
	The whole model can now be described using the block diagram given below.
			\begin{figure}[!htb]
		\begin{center}
		\includegraphics[scale=0.5]{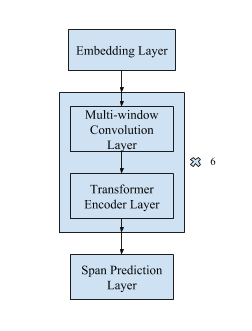}
		\caption{CONV MODEL}
		\end{center}
	\end{figure}
		\end{subsection}
		\end{section}
	
	\begin{section}{BERT\textsubscript{SMALL}}
		A smaller version of the BERT model, with fewer parameters and floating point operations, was also developed and trained for SQuAD task. This model has only 6 Transformer encoder layers\cite{DBLP:journals/corr/VaswaniSPUJGKP17} with hidden size 768, where the BERT\textsubscript{BASE} model has 12. Each $i^{th}$ layer , where $i\in\{1,...,6\}$, of this model was initialized from $(2i-1)^{th}$ layer of the fine tuned BERT\textsubscript{BASE} model. The embedding layer and the span prediction layer were also initialized from fine tuned BERT\textsubscript{BASE}.
		\end{section}
\end{chapter}

%% file: pages/chap.tex
\begin{chapter}{Training}
		\begin{section}{Training BERT}
			BERT\textsubscript{BASE} was fine-tuned for SQuAD task with the default hyperparameters mentioned in the official BERT Github repository. After training for 2 epochs, the model achieved an F1 score of 88.32 on the dev set of SQuAD dataset.
		\end{section}
		\begin{section}{Training CONV MODEL}
		Knowledge distillation, introduced by Hinton et al.\cite{DBLP:journals/corr/HintonVD15}, is a method for compressing the knowledge of an accurate but cumbersome model into a smaller but computationally faster model. For this study, the fine-tuned BERT model is the cumbersome model and it has been ensured that both the models have the exact same output representation. The loss function of the proposed model is described below,
		\begin{gather}
		L=0.5L^S+0.5L^E\\
		\text{where,}\nonumber\\
		L^S = -(0.1\sum_{i=1}^{384}l_i^S\log(\rho_{i1})+0.9T^2\sum_{i=1}^{384}b_{i1}\log(\rho_{i1}))\\
		L^E = -(0.1\sum_{i=1}^{384}l_i^E\log(\rho_{i2})+0.9T^2\sum_{i=1}^{384}b_{i2}\log(\rho_{i2}))\\
		\text{Here, }l^S\in\{0,1\}^{384}\text{ is the ground truth one-hot label.
		}\nonumber\\
		\rho_i = \frac{e^\frac{z_i}{T}}{\sum_{k=1}^{384}e^{\frac{z_k}{T}}}
		\end{gather}
				In this study, $T=3.5$ has been used to generate a softer probability distribution over 384 indices.
		$b_i$ is also a softer probability distribution generated using the logits of the fine-tuned BERT and the same value of $T$. One can observe that, a very high weight of $0.9$ and a low weight of $0.1$ was used on the cross-entropy loss for the soft targets and hard targets respectively.
		\\ \\
		As the fine-tuned BERT model always predicts the start and end indices of the correct span with very high confidence, much of the knowledge about the learned function resides in the ratios of very small probabilities in the soft targets. This knowledge also does not show up in the hard targets provided by the one-hot labels. Using the soft probability distribution over the classes, generated by a cumbersome model for an sample, has been proven helpful in the training of a smaller model by Hinton et al\cite{DBLP:journals/corr/HintonVD15}. As the gradients produced by the soft target get scaled by $\frac{1}{T^2}$, while taking the partial derivatives of the cross-entropy and softmax function, $T^2$ has been multiplied with the cross-entropy loss of the soft targets so that the contributions from both soft targets and hard targets remain somewhat similar.
		\\
		\\
		In some samples, the answer span might not be inside the included context tokens because of the sliding-window approach described in the Embedding layer section. In that case, $l^S$ and $l^E$ has been initialized with zero vectors and then $l_1^S=l_1^E=1$ has been set. It must be noted that the first token in the sequence is NO\_ANS token.
		\\ \\
		As the stride is less than the window size, the tokens can be repeated in multiple windows and there can be multiple scores for a single token. The score with “maximum context” is taken in this scenario. Here the amount of context for a particular token is defined by the minimum of number tokens in the left and right side of that particular token. The same approach has been taken by Devlin et al.\cite{DBLP:journals/corr/abs-1810-04805} too.
		\\ \\
		The model has been trained with ADAM optimizer\cite{DBLP:journals/corr/KingmaB14} with hyperparameters $\beta_1=0.9, \beta_2=0.999, \epsilon=\num{1e-6}$. All the weight matrices and bias vectors were initialized using Glorot normal initializer\cite{Glorot10understandingthe} and zero vectors respectively. The model was trained for 55 epochs with batch-size 60. $L2$ weight decay was used on all weight matrices. Dropout\cite{JMLR:v15:srivastava14a}, with rate 0.2, was used after every affine transformation in the Transformer encoder layer. The learning rate is a function of the current training step and is defined by,
		\begin{gather}
		num\_train\_step = \left\lfloor{(num\_train\_examples\times\frac{total\_epochs}{batch\_size})}\right\rfloor\\
		num\_warmup\_steps = 0.1\times num\_train\_step\\
		\begin{small}
		lr = 
		\begin{cases}
		\frac{current\_step}{num\_warmup\_steps}\times 0.001, & \text{if } current\_step<num\_warmup\_steps\\
		\num{9.999e-4}(1-\frac{current\_step }{ num\_train\_step})^2+\num{1e-7}, & \text{ otherwise}
		\end{cases}
		\end{small}
		\end{gather}
		
		\end{section}
		\begin{section}{Training BERT\textsubscript{SMALL}}
			Training Training BERT\textsubscript{SMALL} is more or less identical to training the proposed model, other than a few hyperparameters. This model was trained for 40 epochs with batch size 30. As knowledge distillation from the fine-tuned BERT\textsubscript{BASE} model was also done, the loss function is identical to the loss function of the proposed model. The learning rate is a function of the current training step and is defined by,
			\begin{gather}
			num\_train\_step = \left\lfloor{(num\_train\_examples\times\frac{total\_epochs}{batch\_size})}\right\rfloor\\
			num\_warmup\_steps = 0.1\times num\_train\_step\\
			lr = 
			\begin{cases}
			\frac{current\_step}{num\_warmup\_steps}\times \num{3e-6}, & \text{if } current\_step<num\_warmup\_steps\\
			\num{3e-6}\times(1-\frac{current\_step }{ num\_train\_step}), & \text{ otherwise}
			\end{cases}
			\end{gather} 
			\end{section}
\end{chapter}

%% file: pages/chap4.tex
\begin{chapter} {Results and Discussion}
	\begin{table}[!htb]
		\begin{center}
			\begin{tabular}{|p{2in}|c|c|}
				\hline
				\textbf{Model} & \textbf{F1} & \textbf{Inference Time (samples/s)} \\  
				\hline
				CONV MODEL & 79.80 & 774 \\
				\hline
				CONV MODEL (with only one CNN with 7 kernel size) & 78.47 & 829 \\
				\hline
				CONV MODEL (no knowledge distillation) & 72.33 & 774 \\
				\hline
				BERT\textsubscript{SMALL} & 85.57 & 217 \\
				\hline
				FABIR & 77.6 & 672 \\
				\hline
				QANet & 82.70 & 163 \\
				\hline
				BiDAF & 77.30 & 60 \\
				\hline
			\end{tabular}
			\caption{F1 and Inference speed of different models}
		\end{center}
	\end{table}
	The proposed two models were developed using Tensorflow library \cite{tensorflow2015-whitepaper} using python language. The experiments were carried out on an NVIDIA v100 GPU. All the inference time measurements were done using $batch size = 100$ on the same hardware. The proposed CONV MODEL was able to achieve faster inference speed and higher F1 score than FABIR. One can notice that the variation of the CONV MODEL with only a single CNN of kernel size 7 yielded less F1 score than the proposed CONV MODEL, which demonstrates the importance of the Multi-Window Convolution layer. One can also observe that without knowledge distillation, the F1 score of the CONV MODEL reduced substantially. It was also observed that without knowledge distillation, the model tends to overfit, which was solved by adding soft targets in the loss function. Correia et al.\cite{DBLP:journals/corr/abs-1810-09580} also report that the architecture proposed by Vaswani et al.\cite{DBLP:journals/corr/VaswaniSPUJGKP17} is more susceptible to overfitting than RNNs. BERT\textsubscript{SMALL} model, consisting of just 6 layers as opposed to BERT\textsubscript{BASE} which consists of 12 Transformer encoder layers, manages to achieve an F1 score of 85.57 while slashing the computations required by BERT\textsubscript{BASE} model approximately by half. It should be noticed that, even with only 6 Transformer encoder layers from fine-tuned BERT\textsubscript{BASE} model, BERT\textsubscript{SMALL} model was able to recover and achieve an F1 score close to its parent(88.32).
	
\end{chapter}

%% file: pages/chap5.tex
\begin{chapter} {Conclusion}

	This study aims at developing a model for MRC task which is computationally fast and performs well. Two models, the CONV MODEL and BERT\textsubscript{SMALL}, have been developed with the goal of achieving this. Experiments show that CONV MODEL is the fastest among the tested models and achieves a reasonably good F1 score whereas BERT\textsubscript{SMALL} model has the best F1 score among all the discussed models. It was also found that Multi-window Convolution Layer improved the F1 score over a single window Convolution layer. Knowledge distillation method increased the F1 score considerably, without the need for a higher capacity model. In future, the same pipeline can be tested for other NLP tasks to test the capability and to improve the model.

\end{chapter}